\documentclass{article}

\usepackage{iclr2027_conference,times}
\usepackage{graphicx}
\usepackage{booktabs}
\usepackage{multirow}
\usepackage{array}
\usepackage{tabularx}
\usepackage{algorithm}
\usepackage{algpseudocode}
\usepackage{xcolor}
\usepackage{amsmath}
\usepackage{amssymb}
\usepackage{float}
\usepackage{wrapfig}
\usepackage{tikz}
\usetikzlibrary{positioning,arrows.meta}
\usepackage{hyperref}
\hypersetup{hidelinks}
\usepackage{url}
\usepackage{enumitem}
\usepackage{listings}
\usepackage{mdframed}
\usepackage{needspace}
\usepackage{etoolbox}
\usepackage{colortbl}

\AtBeginEnvironment{table}{\setlength{\belowcaptionskip}{10pt}}
\AtBeginEnvironment{table*}{\setlength{\belowcaptionskip}{10pt}}

\usepackage{amsmath,amsfonts,bm}

\def\figref#1{figure~\ref{#1}}
\def\Figref#1{Figure~\ref{#1}}

\def\Secref#1{Section~\ref{#1}}

\def\eqref#1{equation~\ref{#1}}
\def\Eqref#1{Equation~\ref{#1}}

\def\algref#1{algorithm~\ref{#1}}
\def\Algref#1{Algorithm~\ref{#1}}

\def\1{\bm{1}}

\DeclareMathAlphabet{\mathsfit}{\encodingdefault}{\sfdefault}{m}{sl}
\SetMathAlphabet{\mathsfit}{bold}{\encodingdefault}{\sfdefault}{bx}{n}

\providecommand{\tabref}[1]{Table~\ref{#1}}
\providecommand{\Tabref}[1]{Table~\ref{#1}}

\providecommand{\Secref}[1]{Section~\ref{#1}}
\providecommand{\figref}[1]{Figure~\ref{#1}}
\providecommand{\Figref}[1]{Figure~\ref{#1}}
\providecommand{\appref}[1]{Appendix~\ref{#1}}
\providecommand{\Eqref}[1]{Equation~\ref{#1}}
\newcommand{\methodname}{BFR}
\newcommand{\fca}{FCA-MS}

\newcommand{\textpolicy}{BFR-Text}
\newcommand{\mvpolicy}{BFR-MV}
\newcommand{\statepolicy}{BFR-State}
\newcommand{\adaptivepolicy}{BFR-Adaptive}

\lstdefinestyle{json}{
  basicstyle=\ttfamily,
  breaklines=true,
  columns=fullflexible,
  showstringspaces=false,
  frame=none,
  xleftmargin=0pt
}

\newenvironment{promptbox}[1]{%
  \begin{mdframed}[
    linewidth=0.8pt,
    linecolor=black!55,
    backgroundcolor=black!4,
    innertopmargin=8pt,
    innerbottommargin=8pt,
    innerleftmargin=10pt,
    innerrightmargin=10pt,
    frametitle={\sffamily\bfseries\color{white}~#1},
    frametitlebackgroundcolor=black!70,
    frametitlerule=false,
    roundcorner=3pt
  ]
}{%
  \end{mdframed}
}

\iclrfinalcopy
\title{Relevance Is Not Sufficiency: What Actually Closes the Evidence Gap in Long-Term Memory QA}

\author{Yufeng Li\textsuperscript{1}\thanks{Yufeng Li and Shuxin Li contributed equally. Emails: \texttt{yufeng03@stu.ecnu.edu.cn} and \texttt{shuxin.li@ntu.edu.sg}.},\enspace Shuxin Li\textsuperscript{2}\footnotemark[1],\enspace Zhenhua Xu\textsuperscript{3},\enspace Junxian Li\textsuperscript{4}
\bfseries Peng Zeng\textsuperscript{1}\thanks{Corresponding author. Email: \texttt{pzeng@sei.ecnu.edu.cn}.},\enspace Sheng Yao\textsuperscript{5}, \\ \bfseries Changting Lin\textsuperscript{3}, 
\bfseries Gaolei Li\textsuperscript{4},\enspace Ran Bi\textsuperscript{6},\enspace Meng Han\textsuperscript{3}\\[0.8ex]
{\normalfont\small \textsuperscript{1}East China Normal University, Shanghai, China}\\
{\normalfont\small \textsuperscript{2}Nanyang Technological University, Singapore}\\
{\normalfont\small \textsuperscript{3}Zhejiang University, Hangzhou, China}\\
{\normalfont\small \textsuperscript{4}Shanghai Jiao Tong University, Shanghai, China}\\
{\normalfont\small \textsuperscript{5}University of Southern California, Los Angeles, USA}\\
{\normalfont\small \textsuperscript{6}Northeastern University, Shenyang, China}}

\begin{document}

\maketitle

\begin{abstract}
LLM agents that interact with a user across many sessions accumulate histories that exceed their context window, so they store past interactions in an external memory and answer each question from a small set of retrieved records. Existing memory systems rank records by lexical or embedding relevance, yet the top-ranked memories can each be relevant while jointly omitting a complementary fact that the answer requires, especially for multi-session and temporal questions. 
Drawing on the distinction between relevance and sufficiency in legal evidence scholarship, we recast memory retrieval as constructing a \emph{sufficient memory set}. 
To operationalize this view, we introduce a blinded LLM judgment over the retrieved set, together with Gold Hit and Turn Hit as evidence-coverage proxies. We then propose \emph{Budgeted Flat Reconstruction (\methodname{})}, which builds sufficient sets over a fixed flat memory store in two stages. Specifically, we first apply \emph{Formal Concept Analysis for Memory Selection (\fca{})} to decompose the question into information requirements and select a compact candidate subset that jointly covers them. Then, we repeatedly acquire unseen records through deeper text search or complementary entity and session views, stopping when the budget is exhausted. 
Experiments on \textsc{LoCoMo} and \textsc{LongMemEval}-S show that \methodname{} outperforms same-store adaptations of recent agent-memory systems in both answer quality and evidence coverage. Specifically, on \textsc{LongMemEval}-S it raises judged accuracy from 72.4\% to 82.2\% and Turn Hit to 91.4\%.
Code is available at: \href{https://github.com/LYF199903/BFR-An-Agent-Memory-Framework-for-Evidence-Reconstruction}{\textbf{GitHub}}.

\end{abstract}

\section{Introduction}

LLM agents interact with users across many sessions, accumulating preferences, events, and other information that may matter much later~\citep{zhong2024memorybank, maharana2024evaluating, DBLP:conf/iclr/WuWYZCY25, tan2025prospect}. Because the full interaction history is too long to remain in the context window~\citep{packer2023memgpt}, these agents commonly rely on external memory to retain and access past information. Agent memory has therefore become an important component for maintaining continuity across conversations and answering questions about earlier interactions. 
Recent agent-memory systems improve answer quality by designing how memories are represented and retrieved.
For example, Mem0 distills conversations into salient facts, consolidates them over time, and retrieves them by embedding similarity~\citep{chhikara2025mem0}. A-Mem instead maintains evolving, interlinked notes retrieved in the same way~\citep{xu2025amem}. CoM builds coherent memory chains from relevance-ranked candidates~\citep{xu2026com}, while MRAgent reasons over a Cue--Tag--Content graph~\citep{ji2026mragent,hu2025memory,luo2026storage}. 
Despite their different memory structures, these systems typically acquire candidates by relevance, whether through lexical matching, embedding similarity, or guided traversal~\citep{zhang2025survey}. 
Yet relevance is a property of each record, not of the set, i.e., a ranking can return several memories about the right topic while the one partner fact the answer requires falls below the cutoff. It means that \textbf{retrieving relevant memories does not by itself ensure that the resulting set can support the answer.}
This raises a question: what determines whether the retrieved records made available to an agent support a correct answer?

\begin{figure*}[t]
\centering
\includegraphics[width=0.8\textwidth]{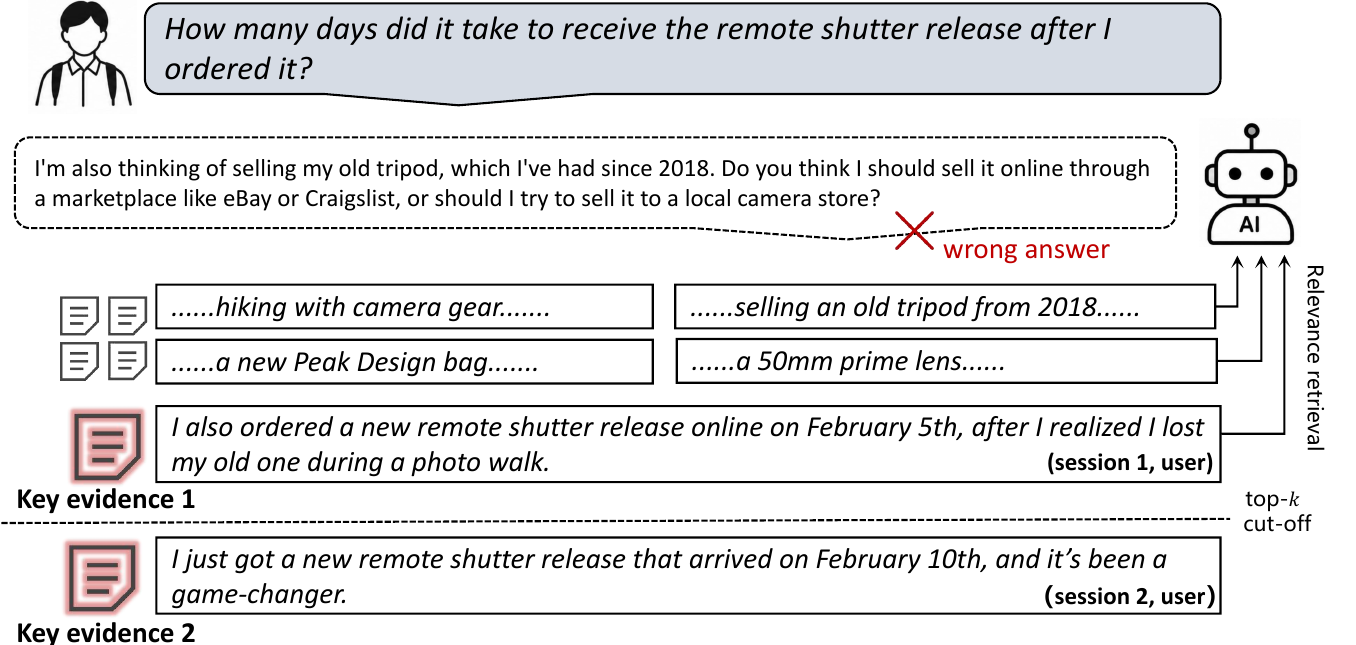}
\caption{\textbf{A real error: relevant but insufficient records.}
The question asks how many days the remote shutter release took to arrive.
Relevance ranking returns Key evidence 1 together with other photography memories, while Key evidence 2 falls below the top-$k$ cutoff.}
\label{fig:overview}
\end{figure*}

Legal evidence scholarship offers a useful distinction: \emph{relevance} concerns whether an individual item bears on a claim, whereas \emph{sufficiency} concerns whether the evidence considered together supports a finding~\citep{lempert1986new, pardo2023evidence}. We apply this distinction to memory QA and ask whether the retrieved memories, taken together, contain enough information to determine the answer. 
\Figref{fig:overview} illustrates this failure with a real \textsc{LongMemEval}-S case. The question asks how many days a remote shutter release took to arrive, which requires two records together: the February 5 order and the February 10 arrival. The top-$k$ retrieved memories include the order alongside other photography-related records, but the arrival, a less similar record from another session, falls below the cutoff. The agent then responds about selling an old tripod instead of the five-day interval. This example shows that individually relevant records do not guarantee a sufficient set.

We therefore recast memory retrieval as the construction of a \textit{sufficient memory set}, one from which the answer can be determined. Constructing such a set raises two coupled challenges. \textit{First, a set must bring together complementary facts.} A question may depend on several pieces of information, and a limited context budget can be spent on records that repeat one piece while leaving another uncovered. \textit{Second, the necessary facts must be reachable.} Evidence missing from the initial candidate pool cannot be recovered by improving selection within that pool. 
Both challenges must be met without knowing which records are required. At inference time, no oracle declares a set sufficient, so a method must itself decide how far to search and when to stop.

To address both challenges, we propose \emph{Budgeted Flat Reconstruction} (\textbf{\methodname{}}), a two-stage, requirement-aware framework that assembles answer-supporting evidence over a flat memory store under a fixed retrieval budget. Specifically, at Stage I, \emph{Formal Concept Analysis for Memory Selection} (\fca{}) first
decomposes the question into answer requirements, and then applies an FCA-based prefilter followed by LLM selection of a compact, complementary subset. These selected candidates are grounded to source records to form the initial evidence set. At Stage II, we acquire further supporting evidence from the current evidence set, excluding records already obtained and accumulating new support within a fixed retrieval budget. We instantiate this stage with two schedules: \emph{Budgeted Flat Reconstruction with Text Access} (\textpolicy{}) extends text retrieval, and \emph{Budgeted Flat Reconstruction with Multi-View Access} (\mvpolicy{}) combines text, entity, and session views. Stage I thus composes the evidence already present among the candidates, Stage II reaches evidence beyond them, and the fixed budget supplies a stopping rule that does not presume sufficiency can be recognized. Finally, we evaluate \methodname{} on \textsc{LoCoMo} and \textsc{LongMemEval}-S under protocols that hold the memory store, answerer, and judge fixed within each benchmark. 
Extensive results show that BFR improves judged answer quality and annotated evidence coverage over same-store baselines. On \textsc{LongMemEval}-S, it raises judged accuracy from $72.4\%$ to $82.2\%$ and attains a $91.4\%$ Turn Hit. In summary, this paper makes three contributions:

(i) We formulate memory retrieval as the construction of a sufficient memory set and make set-level sufficiency measurable: LLM-Suff@Set judges whether a retrieved set supports the answer, while Gold Hit and Turn Hit track coverage of annotated answer-bearing evidence. 

(ii) We propose \methodname{}, a two-stage method for building sufficient sets over a flat memory. Its two stages recover complementary evidence missing within and beyond the initial candidates.

(iii) We conduct extensive experiments on \textsc{LoCoMo} and \textsc{LongMemEval}-S. These results show that \methodname{} improves answer quality and evidence coverage, and identify where additional evidence is recovered and clarify the distinct roles of evidence composition and acquisition.
\section{Related Work}
\label{sec:related}

\textbf{Evidence relevance and sufficiency.}
Evidence scholarship treats relevance as a property of an item's bearing on a claim and sufficiency as a judgment over the complete evidentiary record~\citep{pardo2023evidence}.
In RAG, sufficient-context evaluation uses this distinction diagnostically, separating failures caused by missing context
from failures to use available context~\citep{joren2024sufficient}.
We bring this distinction to long-term agent memory and make it constructive: rather than diagnosing whether a retrieved set was sufficient, retrieval should assemble and complete the evidence set that supports an answer.
\appref{app:pardo} spells out the mapping from Pardo's record-level test to our method.

\textbf{Memory construction and relevance-based access.}
Memory systems increasingly transform interaction history before retrieval.
Mem0 maintains salient facts through add, update, and delete operations and retrieves
semantically similar memories~\citep{chhikara2025mem0}.
A-Mem creates evolving linked notes but still answers from a top-$k$
similarity-ranked subset~\citep{xu2025amem}.
Zep represents temporal validity in a bi-temporal graph
~\citep{rasmussen2025zep}; MemoryOS uses hierarchical storage
~\citep{kang2025memoryos}; and SeCom segments conversations into coherent units
~\citep{pan2025secom}.
CoM improves post-retrieval organization by growing chains from top-ranked anchors
according to query relevance and contextual consistency~\citep{xu2026com}.
MRAgent applies active reconstruction on a Cue--Tag--Content memory graph, allowing
an LLM to choose retrieval tools from accumulated evidence
~\citep{ji2026mragent}.
These systems improve what can be retrieved or how relevant candidates are organized.
\methodname{} instead asks whether the resulting set covers all answer requirements,
then acquires missing support without changing the underlying store.

\section{WHAT COUNTS AS A SUFFICIENT CANDIDATE MEMORY SET}
\label{sec:problem}
\label{sec:setting}

\begin{figure*}[t]
\centering
\includegraphics[width=\textwidth]{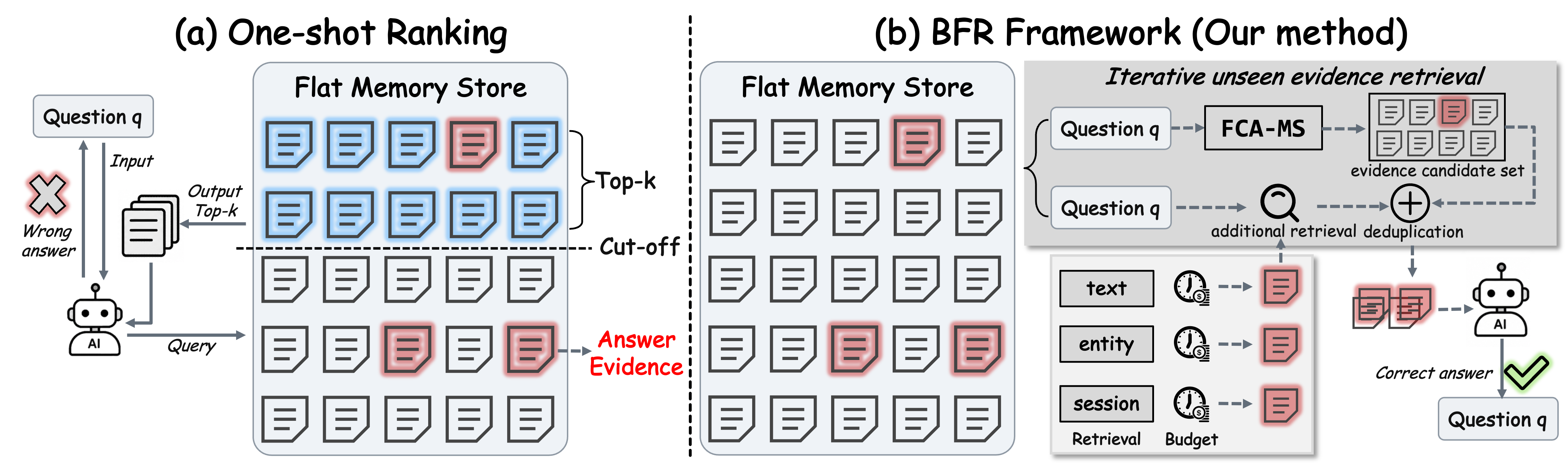}
\caption{\textbf{Similarity top-$k$ vs. set-level sufficiency.}
(a)~Prior memory architectures rank records by query similarity and return the top-$k$
slice.
(b)~\methodname{} returns a sufficient set over the flat store: it first selects complementary records that jointly cover the question's requirements, then completes missing support under a budget by adding unseen records through text, entity, and session views.
}
\vspace{-10pt}
\label{fig:arch}
\end{figure*}

\textbf{Motivation.}\label{sec:motivation}
Similarity search can return several memories about the right topic while missing
the fact that makes an answer possible.
In \figref{fig:overview}, the retrieved records include the shutter-release order
and other photography-related memories, but not the arrival date.
The records are relevant to the question; together they cannot establish the
delivery interval.
Legal evidence scholarship makes the same item-versus-set distinction:
an item can bear on a claim without the assembled record being sufficient to
support a finding~\citep{lempert1986new,pardo2023evidence} (\appref{app:pardo}).
For memory QA, the practical difficulty is twofold.
Relevant candidates may repeat one part of the answer instead of supplying
complementary facts, and a necessary fact may fall outside the initial candidate
pool even when it remains in the memory store.
\Figref{fig:arch} illustrates why a memory system must address both the composition
of the retrieved set and access to evidence beyond the first retrieval boundary.

\textbf{Quantifying candidate-set sufficiency.}\label{sec:suff_eval}
Let $q$ be a question, $y^\star(q)$ its reference answer, $\mathcal{M}$ a fixed
memory store, and $E\subseteq\mathcal{M}$ the retrieved candidate set. We
write $\operatorname{text}(E)$ for the evidence text rendered from $E$ and
$\operatorname{src}(E)$ for the set of source-record or source-turn identifiers
linked to $E$. As usual, $\mathbb{1}[P]$ equals $1$ when proposition $P$ holds
and $0$ otherwise. The ideal set-level property is
\begin{equation}
\label{eq:sufficiency}
S(q,E)=\mathbb{1}\!\left[y^\star(q)\text{ is determinable from }
q\text{ and }\operatorname{text}(E)\right].
\end{equation}
Thus, $S(q,E)$ concerns what the evidence makes answerable, independently of
whether a particular answerer generates the correct response.
To assess this property offline, let $a(q,E)$ be the final \emph{decided} label
from a blinded, reference-conditioned LLM adjudication given $q$,
$y^\star(q)$, and $\operatorname{text}(E)$ (\appref{app:annotation}), with
$a(q,E)\in\{\mathrm{sufficient},\mathrm{insufficient}\}$.  We define
\emph{LLM-Suff@Set} by
\begin{equation}
\label{eq:llm_suffset}
\widehat S_{\mathrm{LLM}}(q,E)
=\mathbb{1}\!\left[a(q,E)=\mathrm{sufficient}\right].
\end{equation}
Here, $\widehat S_{\mathrm{LLM}}(q,E)$ is the operational, reference-conditioned
assessment of joint answer support. It is undefined when adjudication is
uncertain or missing; such cases are excluded from the reported rate rather
than counted as insufficient.
For questions with annotated supporting records, let $\mathcal{G}(q)$ be their
nonempty set of source identifiers and $\mathcal{H}(q)$ be their set of annotated source-turn identifiers, which may be empty. \emph{Gold Hit} and \emph{Turn Hit} are defined as follows: 
\begin{align}
G(q,E) &=\mathbb{1}\!\left[
\operatorname{src}(E)\cap\mathcal{G}(q)\neq\varnothing\right] \label{eq:reach} \\
\operatorname{TH}(q,E)&=\mathbb{1}\!\left[
\operatorname{src}(E)\cap\mathcal{H} (q)\neq\varnothing\right] \label{eq:turn_hit}
\end{align}
The value $G(q,E)$ is $1$ if $E$ reaches at least one annotated supporting
record; it does not require retrieval of every supporting record. Gold Hit is
undefined when $\mathcal G(q)=\varnothing$ and is averaged over questions with
nonempty supporting-record annotations.
The value $\operatorname{TH}(q,E)$ is $1$ if $E$ reaches at least one annotated
answer-bearing turn. Our main \textsc{LongMemEval}-S results average Turn Hit
over all $500$ questions. These proxies measure whether annotated evidence was reached, not whether the set suffices. In \tabref{tab:sufficiency_example}, each singleton has Gold Hit $1$, and the
arrival alone has Turn Hit $1$. Neither yields the five-day answer, but together
they do. A hit is therefore necessary for sufficiency but does not establish it, which is why we report the proxies alongside, not in place of, the set-level judgment.

\begin{table}[t]
\centering
\caption{\textbf{Illustrative delivery-time question.}
Both dates are required to derive the five-day answer.}
\label{tab:sufficiency_example}
\begin{tabular}{lccc}
\toprule
Retrieved evidence & LLM-Suff@Set & Gold Hit & Turn Hit \\
\midrule
\textit{Order only (Feb.~5)} & 0 & 1 & 0 \\
\textit{Arrival only (Feb.~10)} & 0 & 1 & 1 \\
\textbf{Order + arrival} & \textbf{1} & 1 & 1 \\
\bottomrule
\end{tabular}
\vspace{-8pt}
\end{table}

\section{\methodname: Composing and Completing Sufficient Evidence}
\label{sec:method}

\subsection{Problem: construct a sufficient set under a retrieval budget}
\label{sec:bfr_overview}

Let $\mathcal{M}=\{m_1,\dots,m_N\}$ be a fixed memory store, $q$ a question,
and $y^\star$ its reference answer.
An access policy constructs $E\subseteq\mathcal{M}$, after which the fixed
answerer $f$ emits $\hat{y}=f(q,E)$.
The set-level sufficiency property $S(q,E)$ is defined in \Secref{sec:problem}.
Holding $\mathcal{M}$ and $f$ fixed, the construction problem is to maximize
expected sufficiency under a retrieval-call budget $B$:
\begin{equation}
\label{eq:obj}
\max_{\pi}\ \mathbb{E}\!\left[S(q,E_T)\right]
\quad \text{s.t.}\quad C_{\mathrm{ret}}(\pi,q)\leq B .
\end{equation}
A one-shot reader returns $E=\operatorname{TopK}(q,\mathcal{M})$ and cannot recover
once a required exhibit is missing.
Budgeted reconstruction instead starts from an initial set $E_0$ and adds only
unseen records,
\begin{equation}
\label{eq:update}
E_{t+1}=E_t\cup K_t,\qquad
K_t\subseteq \mathcal{M}\setminus E_t .
\end{equation}
\Eqref{eq:obj} is the problem. The two gaps in \Secref{sec:motivation} say
what a solution must do.
Composition chooses $E_0$ so that distinct answer requirements are covered rather
than redundant relevant items.
Completion chooses each $K_t$ so that a missing exhibit can still enter the record
after the first page.

\begin{figure*}[t]
\centering
\includegraphics[width=\textwidth]{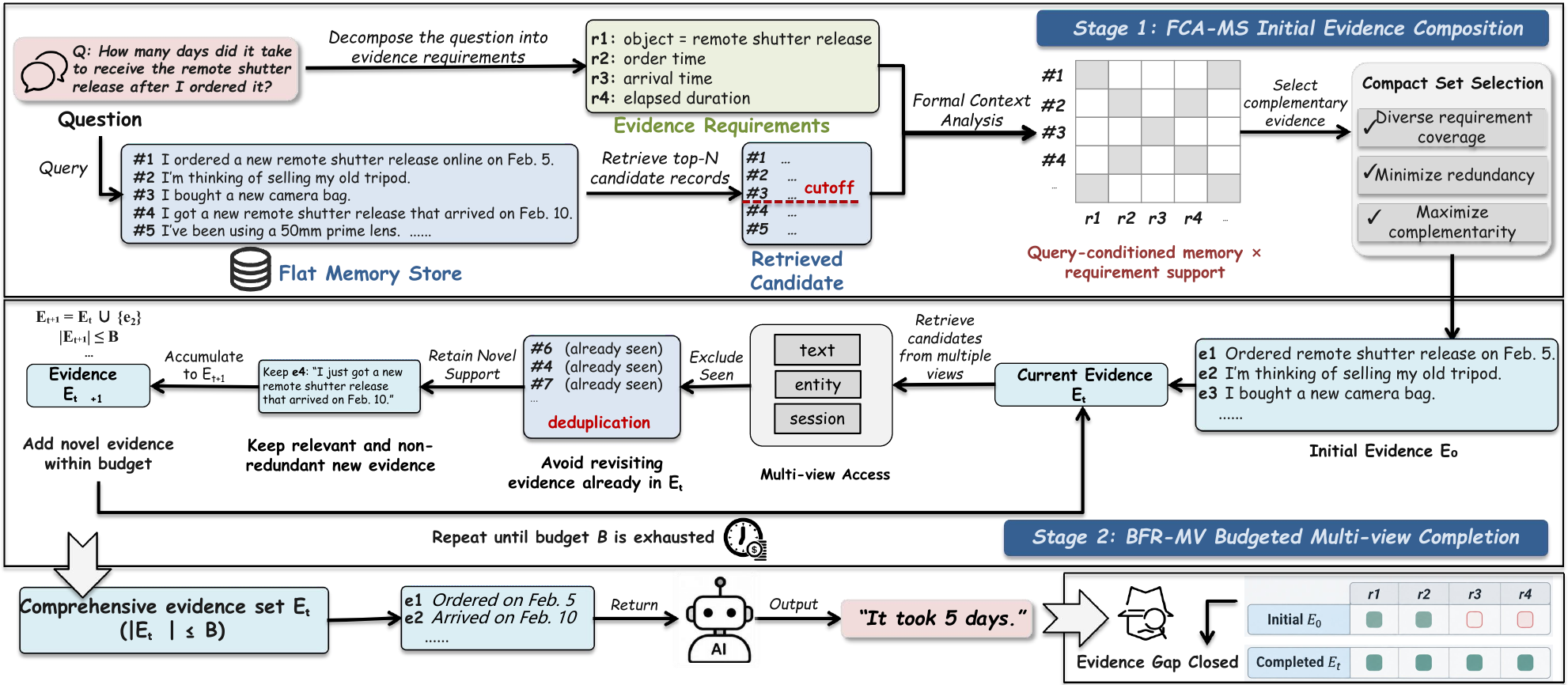}
\vspace{-5pt}
\caption{\textbf{The two-stage evidence-construction pipeline in \methodname{}.}
Stage~I (\fca{}) retrieves a candidate pool larger than the usual top-$k$, decomposes the query into answer requirements, prefilters candidates with FCA, and has an LLM select a compact subset that jointly covers the requirements. The selected candidates are mapped to source records to form $E_0$.
Stage~II (shown for \mvpolicy{} instantiation) completes the set under a fixed budget: text, entity, and session access retrieve unseen records. Only new records are accumulated into $E_{t+1}$ until the budget is exhausted. The comprehensive evidence set returns to Agent, and Agent answers the question based on the set.}
\label{fig:bfr_pipeline}
\vspace{-10pt}
\end{figure*}

\methodname{} assigns one stage to each gap.
Let $\widetilde{\mathcal{M}}$ be an agent-memory note representation and
$\phi$ a grounding map into $\mathcal{M}$.
Stage I builds $E_0$ with \fca{} and Stage II runs a completion schedule $\pi$:
\begin{equation}
\begin{aligned}
E_0 &= \phi\!\left(\rho_{\mathrm{FCA}}(q,\widetilde{\mathcal{M}})\right),\\
A_t &= \pi(q,E_t,t),\\
E_{t+1} &= E_t \cup \kappa_\pi(q,E_t,C_t),
\qquad t<T.
\end{aligned}
\label{eq:bfr_framework}
\end{equation}
$A_t$ names the retrieval views and cues at round $t$, $C_t$ is the unseen
candidate pool those calls return, and $\kappa_\pi$ retains added records.
The answerer reads $E_T$ after the budget is spent.
\Figref{fig:bfr_pipeline} makes this data flow explicit.
Stage~I asks the agent memory system for an expanded candidate pool beyond its
usual top-$k$, decomposes the query into requirements, filters the pool by
requirement-aware FCA, and asks an LLM to select a compact supporting subset.
A bounded grounding map then supplies the
initial source-record evidence $E_0$.
Stage~II treats this set as a cumulative evidence state, retrieves through the
views specified by $\pi$, removes records already in $E_t$, and retains novel
support until an observable budget is exhausted.
The figure instantiates this loop with \mvpolicy{}; \textpolicy{} uses the same
exclusion, retention, accumulation, and stopping operations with text access only.
\appref{app:pardo} records how $S$, $R$, $\mathcal{R}_q$, and the two stages
correspond to Pardo's item/record distinction.

\subsection{Stage I: Compose complementary evidence with \fca{}}
\label{sec:fca_ms}

\textbf{Expanded retrieval and requirement decomposition.}
An agent memory system returns an ordered pool
$\mathcal{C}_q=\operatorname{TopN}(\operatorname{Retrieve}_{\mathrm{agent}}
(q,\widetilde{\mathcal{M}}))$, with $N>k$ relative to its usual top-$k$
retrieval cutoff.  An LLM decomposes $q$ into answer-oriented
requirements $\mathcal{R}_q=\{r_j\}_{j=1}^{J}$, including distinct facts, bridges,
list items, and temporal constraints.

\textbf{Requirement-aware prefilter.}
An FCA-based prefilter uses the requirements and retrieved candidates to retain
a smaller pool $\mathcal P_q=\operatorname{FCAPrefilter}(\mathcal R_q,\mathcal C_q)$.
This step exposes a broader range of potentially complementary notes to the
selector while bounding its input.

\textbf{Minimum-subset selection and grounding.}
An LLM receives $q$ and $\mathcal P_q$ and is instructed to select the smallest
memory subset that jointly supplies the required information.  Its selected IDs
are checked against the prefiltered pool and deduplicated; the implementation
uses requirement-coverage output and a fixed fallback if the valid subset is
empty (\appref{app:impl}).  Only the resulting selected notes $S$ are grounded
to source turns by $\phi$ to produce $E_0\subseteq\mathcal M$; the retriever's
initial top-$k$ is a reference cutoff, not a retained base set.

This stage improves requirement coverage within the initial candidate pool but remains bounded by that pool. A required record can therefore be absent from $E_0$ even when $E_0$ already contains several records related to the question. 
Stage II treats this remaining gap as an acquisition problem instead of reselecting repeatedly from the
same pool. Algorithm~\ref{alg:fca_ms} in \appref{app:algorithms} summarizes the
selection and grounding steps.

\subsection{Stage II: Complete missing support under a budget}
\label{sec:bfr_loop}

\textbf{Multi-view access to one fixed store.}
Text, entity, and session indices provide different access paths to the same
source-grounded records in $\mathcal{M}$.  At round $t$, each action
$a=(v,c)\in A_t$ specifies a view $v$ and a cue $c$.  The retrievers exclude all
records already contained in $E_t$:
\begin{equation}
C_t=\operatorname{Unique}\!\left(
\bigcup_{(v,c)\in A_t}R_v(c,\mathcal{M}\setminus E_t)\right).
\label{eq:candidates}
\end{equation}
The exclusion constraint makes acquisition cumulative and duplicate-free. The
retention rule $\kappa_\pi$ ranks the returned records by relevance and by novelty with respect to the current set, keeps at most $k_{\mathrm{keep}}$ of them, and adds them to $E_t$ by set union.  Consequently,
each round either extends the reachable evidence or leaves the set unchanged.

\textbf{Budgeted stopping.}
The budget
$B=(B_{\mathrm{step}},B_{\mathrm{call}},B_{\mathrm{new}})$ limits the number of
rounds, retrieval calls, and newly retained records.  The main configuration allows
three rounds, six calls, and $30$ new records.  We use budget exhaustion as the
stopping condition because the offline sufficiency measures in
\Secref{sec:suff_eval} require either a reference answer or gold annotations and are
therefore unavailable at inference time.
Algorithm~\ref{alg:bfr_completion} in \appref{app:algorithms} gives the shared
completion loop; the fixed schedules differ only in how they instantiate $A_t$.

\subsection{Access schedules for completion}
\label{sec:bfr_policies}

The shared completion loop admits two fixed schedules for the two ways required
support can remain outside $E_0$.
They share \fca{}, $\mathcal{M}$, the exclusion rule, the budget, and the answerer;
only the allocation of retrieval calls changes. We also evaluate state-conditioned completion as a controlled variant of Stage~II;
its settings and results are reported in \Secref{sec:mechanism} and
\appref{app:budget}.

\textbf{\textpolicy{}: deepen the text ranking.}
Each round queries the text index with the original question $q$, removes ids already
in $E_t$, and keeps the first $k_{\mathrm{keep}}=8$ unseen hits.  Repeating the same
query advances beyond the initial page without query rewriting.  This schedule
targets cases in which the missing evidence is relevant to $q$ but appears below
the first cutoff.

\textbf{\mvpolicy{}: diversify the access path.}
The multi-view schedule issues text, entity, and session actions in a fixed order.
The text action uses $q$; the entity action uses the first person named in $q$ when
available; the session action uses the latest retained record as an anchor.
Entity and session access can recover records that do not rank highly under the full
question string.

\section{Experiments}
\label{sec:experiments}

In this section, we conduct controlled experiments to evaluate \methodname{} and
answer four research questions:
\textbf{(RQ1)} How does \methodname{} compare with same-store memory baselines in
answer quality?
\textbf{(RQ2)} Do retrieved sets contain sufficient evidence to answer the
question, and how does direct sufficiency compare with annotated-support hits?
\textbf{(RQ3)} How do the two stages and access schedules contribute to
performance?
\textbf{(RQ4)} How do answer quality and evidence coverage change with
additional retrieval cost?
The appendix reports full operating-point, question-type, and cross-system
results.

\subsection{Experimental setup}
\label{sec:setup}

\paragraph{Datasets.}
We evaluate on two benchmarks for long-term conversational memory.
\textsc{LoCoMo}~\citep{maharana2024evaluating} tests understanding of long,
multi-session conversations, including questions that require linking information
across turns.
\textsc{LongMemEval}-S~\citep{DBLP:conf/iclr/WuWYZCY25} tests whether a memory system
can answer questions about information distributed across extended interaction
histories and sessions.
Both provide evidence annotations, allowing us to assess retrieval as well as
final answers.
To keep multi-arm evaluation costs manageable, we use a fixed stratified
$200$-question \textsc{LoCoMo} development subset; we evaluate all $500$ questions
of \textsc{LongMemEval}-S.
Dataset provenance and the construction of every derived evaluation subset are
documented in \appref{app:data_provenance}.

\paragraph{Baselines.}
We evaluate \methodname{} against four representative memory-augmented baselines:
Mem0~\citep{chhikara2025mem0}, A-Mem~\citep{xu2025amem},
CoM$^*$~\citep{xu2026com}, and MRAgent-Flat$^*$~\citep{ji2026mragent}.
Within each benchmark, all arms share the same flat turn store, extractive
answerer, and answer judge. The starred methods are controlled same-store
adaptations of the published systems. Baseline retrieval and \methodname{}
initialization are detailed in \appref{app:impl}.

\textbf{Metrics.}
Answer quality is measured by Porter stem F1 and LLM-Judge (J) on \textsc{LoCoMo},
and by Judge on \textsc{LongMemEval}-S.  The answer judge is
\texttt{gpt-5.5} at temperature zero with prompt \texttt{mem0\_accuracy\_v1}.
On \textsc{LoCoMo}, $\mathcal{G}(q)$ is the released set of supporting-record
identifiers.  We report the sufficiency indicators in \Secref{sec:suff_eval}
and two finer measures of annotated-evidence recovery:
\begin{equation}
\operatorname{ER}(q,E)=
\frac{|\operatorname{src}(E)\cap\mathcal{G}(q)|}{|\mathcal{G}(q)|},\qquad
\operatorname{CER}(q,E)=
\mathbb{1}[\mathcal{G}(q)\subseteq\operatorname{src}(E)].
\label{eq:evidence_recall}
\end{equation}
Evidence Recall (ER) is the fraction of annotated supporting records in $\mathcal{G}(q)$ that $E$ recovers, while Complete Evidence Recall (CER) requires every annotated supporting record.
On \textsc{LongMemEval}-S, $\mathcal{H}(q)$ comprises turns marked
\texttt{has\_answer}; Turn Hit measures whether the retrieved set reaches one.
These retrieval and set-level measures complement rather than
replace end-to-end answer scores.
Prompts are listed in \appref{app:prompts}.

\subsection{Experimental Result}

\textbf{RQ1: Does BFR improve answer quality under the same memory store?}
\label{sec:main}

\begin{table*}[t]
\centering
\setlength{\tabcolsep}{3.2pt}
\caption{\textbf{Performance across different question types on \textsc{LoCoMo}}
($n{=}200$).
Same flat store and extractive answerer; BFR is BFR-MV C6.
F1 and LLM-Judge (J) are percentages. Bold denotes the best value in each column.
}
\label{tab:main_locomo}
\begin{tabular}{l cc cc cc cc cc}
\toprule
 \multirow{2}{*}{Method} & \multicolumn{2}{c}{Multi-Hop} & \multicolumn{2}{c}{Temporal} & \multicolumn{2}{c}{Open Domain} & \multicolumn{2}{c}{Single-Hop} & \multicolumn{2}{c}{Overall} \\
\cmidrule(lr){2-3}\cmidrule(lr){4-5}\cmidrule(lr){6-7}\cmidrule(lr){8-9}\cmidrule(lr){10-11}
 & F1 & J & F1 & J & F1 & J & F1 & J & F1 & J \\
\midrule
  Mem0~\citep{chhikara2025mem0} & 29.3 & 42.4 & 74.1 & 78.0 & 2.2 & 0.0 & 58.5 & 59.6 & 55.3 & 59.0 \\
  A-Mem~\citep{xu2025amem} & 32.3 & 42.4 & 70.2 & 72.0 & \textbf{3.8} & 0.0 & 47.1 & 51.4 & 48.7 & 53.0 \\
  CoM$^*$~\citep{xu2026com} & 29.3 & 36.4 & 66.2 & 70.0 & 3.2 & 0.0 & 52.3 & 56.0 & 50.0 & 54.0 \\
  MRAgent-Flat$^*$~\citep{ji2026mragent} & 32.8 & 45.5 & 72.1 & 78.0 & 2.9 & 0.0 & 63.2 & 68.8 & 58.0 & 64.5 \\
  \rowcolor{black!8}
  \textbf{BFR (ours)} & \textbf{44.6} & \textbf{48.5} & \textbf{78.1} & \textbf{80.0} & 2.7 & \textbf{12.5} & \textbf{75.7} & \textbf{77.1} & \textbf{68.3} & \textbf{70.5} \\
\bottomrule
\end{tabular}
\end{table*}

\begin{table*}[t]
\centering
\setlength{\tabcolsep}{4pt}
\caption{\textbf{Performance across different question types on
\textsc{LongMemEval}-S} ($n{=}500$).
The four question-type columns report Judge accuracy; All values are percentages.
All rows share the same turn store, extractive answerer, and judge.
Bold and underline denote the best and second-best values in each column.
The full six-type Turn Hit breakdown is in \appref{app:lme_types}.}
\label{tab:main_lme}
\begin{tabular}{lcccccc}
\toprule
 & \multicolumn{4}{c}{Judge by question type} & \multicolumn{2}{c}{Overall} \\
\cmidrule(lr){2-5}\cmidrule(lr){6-7}
Method & \shortstack{Multi-S} & \shortstack{Single-S}
       & Temporal & Preference & Judge & Turn Hit \\
\midrule
Mem0~\citep{chhikara2025mem0} & \underline{60.9} & 87.1 & \underline{65.4} & 26.7 & \underline{72.4} & 64.8 \\
A-Mem~\citep{xu2025amem} & 54.1 & 87.1 & 57.1 & \textbf{33.3} & 68.6 & \underline{82.2} \\
CoM~$^*$\citep{xu2026com} & 54.1 & 87.1 & 57.1 & \textbf{33.3} & 68.6 & 80.6 \\
MRAgent-Flat$^*$~\citep{ji2026mragent} & 58.6 & \underline{90.0} & 63.2 & \underline{30.0} & 71.4 & 78.8 \\
\rowcolor{black!8}
BFR (ours) & \textbf{69.2} & \textbf{92.9} & \textbf{84.2} & \underline{30.0} & \textbf{82.2} & \textbf{91.4} \\
\bottomrule
\end{tabular}
\end{table*}

\textit{\methodname{} achieves the best overall answer quality among same-store methods
on both benchmarks (\tabref{tab:main_locomo}, \tabref{tab:main_lme}).}
On \textsc{LoCoMo}, it leads in Stem F1 and Judge across Multi-Hop, Temporal,
and Single-Hop questions.
On \textsc{LongMemEval}-S, it also leads in overall Judge and Turn Hit, with a
particularly strong gain on Temporal questions.
The small \textsc{LoCoMo} Open Domain slice is reported without a separate
type-level conclusion. These results are consistent with the two-stage design: \fca{} assembles
complementary initial evidence, and budgeted completion retrieves unseen
support beyond the initial set.
The following analyses test whether this support closes evidence gaps and how
stage, access schedule, and budget contribute.
Paired same-store comparisons and the separate comparison with CoM and
Full-MRAgent are in \appref{app:locomo_controls} and \appref{app:graph},
respectively.

\begin{table*}[t]
\centering
\setlength{\tabcolsep}{3.6pt}
\caption{\textbf{Set sufficiency and annotated-evidence reach across same-store
systems.}
On \textsc{LoCoMo} ($n{=}200$), LLM-Suff@Set is the blinded set-level judgment
and Gold Hit measures annotated-evidence access.
On \textsc{LongMemEval}-S, LLM-Suff@Set is the set-level judgment and
Turn Hit measures annotated-evidence access over all $500$ questions,
with unannotated items counted as misses. All values are percentages.
BFR-Text uses C3 and BFR-MV uses C6.}
\label{tab:locomo_suff_systems}
\begin{tabular}{lcccc}
\toprule
\multirow{2}{*}{System} & \multicolumn{2}{c}{\textsc{LoCoMo} ($n{=}200$)}
& \multicolumn{2}{c}{\textsc{LongMemEval}-S ($n{=}500$)} \\
\cmidrule(lr){2-3}\cmidrule(lr){4-5}
& \shortstack{LLM-Suff@Set} & Gold Hit
& \shortstack{LLM-Suff@Set} & Turn Hit \\
\midrule
Mem0~\citep{chhikara2025mem0} & 45.5\% & 57.0\% & 48.0\% & 64.8\% \\
A-Mem~\citep{xu2025amem} & 36.5\% & 50.0\% & 38.0\% & 82.2\% \\
CoM$^*$~\citep{xu2026com} & 44.0\% & 49.0\% & 49.0\% & 80.6\% \\
MRAgent-Flat$^*$~\citep{ji2026mragent} & 54.5\% & 62.5\% & 56.0\% & 78.8\% \\
\rowcolor{black!8}
\fca{} (Stage I Only) & 42.5\% & 51.0\% & 48.0\% & 85.0\% \\
\rowcolor{black!8}
BFR-Text & 62.0\% & 72.5\% & 63.0\% & 88.6\% \\
\rowcolor{black!8}
BFR-MV & \textbf{63.5\%} & \textbf{73.5\%} & \textbf{65.0\%} & \textbf{91.4\%} \\
\bottomrule
\end{tabular}
\end{table*}

\begin{table*}[t]
\centering
\setlength{\tabcolsep}{5pt}
\caption{\textbf{Where budgeted completion closes the evidence gap on
\textsc{LoCoMo}} ($n{=}200$).
CER is Complete Evidence Recall; LLM-Suff@Set is the blinded set-level judgment.
Text and MV denote BFR-Text at C3 and BFR-MV at C6, with mean realized calls of
$3.00$ and $5.99$; FCA-MS uses no completion calls.
Open Domain contains only eight questions and is interpreted descriptively.}
\label{tab:motivation_budget}
\begin{tabular}{l r cc cc cc}
\toprule
\multirow{2}{*}{Question type} & \multirow{2}{*}{$n$} & \multicolumn{2}{c}{FCA-MS} &
\multicolumn{2}{c}{BFR-Text} &
\multicolumn{2}{c}{BFR-MV} \\
\cmidrule(lr){3-4}\cmidrule(lr){5-6}\cmidrule(lr){7-8}
 &  & CER & LLM-Suff@Set & CER & LLM-Suff@Set & CER & LLM-Suff@Set \\
\midrule
Multi-Hop  & 33  & 6.1 & 15.2 & 18.2 & 27.3 & 12.1 & 30.3 \\
Temporal   & 50  & 66.0 & 60.0 & 82.0 & 74.0 & 82.0 & 74.0 \\
Open Domain & 8 & 12.5 & 0.0 & 12.5 & 0.0 & 25.0 & 0.0 \\
Single-Hop & 109 & 41.3 & 45.9 & 71.6 & 71.6 & 70.6 & 73.4 \\
\midrule
\rowcolor{black!8}
Overall & 200 & 40.5 & 42.5 & 63.0 & 62.0 & 62.0 & 63.5 \\
\bottomrule
\end{tabular}
\end{table*}

\textbf{RQ2: Do the retrieved sets contain the evidence needed for an answer?}
\label{sec:evidence_sufficiency}

\begin{wraptable}{r}{0.6\linewidth}
\centering
\vspace{-20pt}
\caption{\textbf{Stage and access-schedule ablation.}}
\label{tab:stage_schedule_ablation}
\begin{tabular}{lcc|cc}
\toprule
\multirow{2}{*}{Configuration} & \multicolumn{2}{c}{\textsc{LoCoMo}} &
\multicolumn{2}{c}{\textsc{LongMemEval}-S} \\
\cmidrule(lr){2-3}\cmidrule(lr){4-5}
 & F1 & $J$ & $J$ & TH \\
\midrule
FCA-MS  & 48.6 & 53.5 & 70.4 & 85.0 \\
FCA-MS \\ \quad \textit{w/ BFR-Text} & 67.8 & 70.0 & 81.6 & 88.6 \\
FCA-MS \\ \quad \textit{w/ BFR-MV}  & \textbf{68.3} & \textbf{70.5} &
\textbf{82.2} & \textbf{91.4} \\
\bottomrule
\end{tabular}
\end{wraptable}
\textit{\methodname{} achieves the highest observed LLM-Suff@Set on both benchmarks among the compared same-store systems  (\tabref{tab:locomo_suff_systems}).}
Under the main budget configuration, both completion schedules exceed \fca{} alone and all baselines on set-level sufficiency and annotated-evidence coverage, with \mvpolicy{} highest on every measure. Since \fca{} alone trails both schedules, the gain comes from completing the initial set beyond its candidate pool rather than from selection within it.

\Tabref{tab:motivation_budget} further shows that completion improves complete recovery of annotated evidence on \textsc{LoCoMo}. However, the two measures rank the schedules differently: \textpolicy{} has higher overall CER, whereas \mvpolicy{} has higher LLM-Suff@Set. On \textsc{LongMemEval}-S, A-Mem likewise has higher Turn Hit than Mem0 but lower set-level sufficiency. Thus, annotated-evidence access alone does not establish whether the retrieved set jointly supports the answer. Detailed \textsc{LoCoMo} ER, CER, and LLM-Suff@Set results are in \appref{app:budget}, and the six-type \textsc{LongMemEval}-S Turn Hit breakdown is in \appref{app:lme_types}.

\textbf{RQ3: Which stage and access schedule produce the gains?}
\label{sec:mechanism}

\begin{wrapfigure}{r}{0.55\textwidth}
  \centering
  \vspace{-10pt}
  \includegraphics[width=0.5\textwidth]{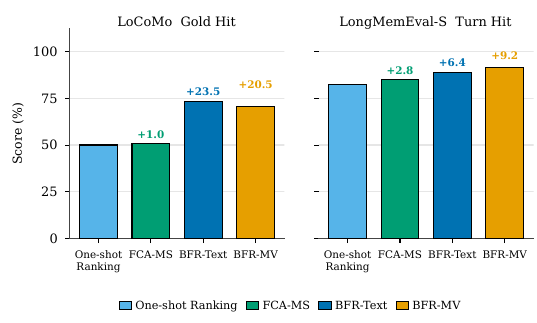}
  \vspace{-5pt}
  \caption{\textbf{Coverage relative to one-shot ranking.}}
  \label{fig:ablation-coverage}
\vspace{-5pt}
\end{wrapfigure}
\fca{} improves annotated-evidence reach over one-shot ranking on both
benchmarks; budgeted completion from this initializer yields further gains in
evidence access and answer quality.
In \Figref{fig:ablation-coverage}, \fca{} raises \textsc{LoCoMo} Gold Hit from
$50.0\%$ to $51.0\%$ and \textsc{LongMemEval}-S Turn Hit from $82.2\%$ to
$85.0\%$. Adding either BFR schedule retrieves more annotated support at the
reported operating points. \Tabref{tab:stage_schedule_ablation} shows that,
relative to \fca{} alone, both schedules also raise Stem F1 and Judge on
\textsc{LoCoMo} and Judge on \textsc{LongMemEval}-S. The schedules favor different reported operating points. At the matched C3
point in \figref{fig:ablation-coverage}, \textpolicy{} has higher
\textsc{LoCoMo} Gold Hit than \mvpolicy{}. The full \mvpolicy{} C6
configuration has the highest \textsc{LongMemEval}-S Judge and Turn Hit in
\tabref{tab:stage_schedule_ablation}, at a higher call budget than
\textpolicy{} C3. Type-level results are in
\tabref{tab:locomo_schedule_types} and \appref{app:lme_types}.

\textbf{RQ4: What does additional retrieval cost buy?}
\label{sec:budget_question}

\begin{wrapfigure}{r}{0.58\linewidth}
\centering
\vspace{-10pt}
\includegraphics[width=\linewidth]{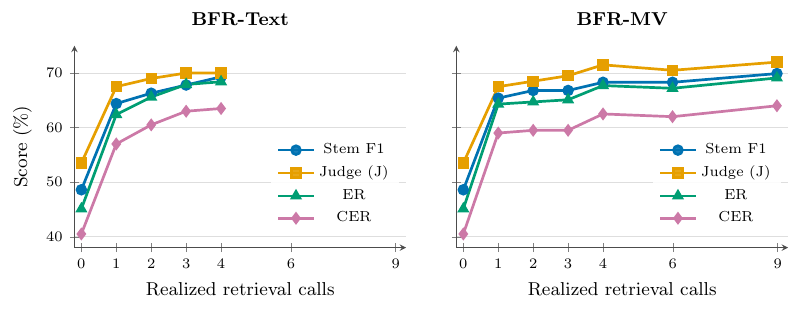}
\vspace{-10pt}
\caption{\textbf{Answer quality and annotated-evidence recovery versus retrieval cost on \textsc{LoCoMo}} ($n{=}200$).}
\vspace{-6pt}
\label{fig:budget_calls}
\end{wrapfigure}
\Figref{fig:budget_calls} shows the cost--quality tradeoff. The panels show BFR-Text and BFR-MV starting from the same FCA-MS C0 set.
The horizontal axis is the mean realized number of retrieval calls; ER and CER
measure partial and complete recovery of annotated evidence, respectively.
Lines connect observed operating points only. BFR-Text C4 and C6 coincide at
four realized calls. Most of the answer-quality gain over FCA-MS appears after the first completion
call. At the matched three-call point, BFR-Text exceeds BFR-MV on all four
plotted metrics. BFR-MV reaches its highest scores at C9, but requires nearly
nine calls, and its curves are not monotonic between budgets.
The C6 configuration in the main tables remains the fixed default operating
point rather than a sweep-selected maximum.
The complete numerical grid, including evidence-set size and LLM-Suff@Set,
is in \appref{app:budget}; the matched-call RSC comparison is in
\appref{app:rsc}.

\section{Conclusion and Discussion}
\label{sec:conclusion}

In this paper, we proposed \methodname{}, a two-stage method for building sufficient evidence sets that can support answers in long-term memory QA. On \textsc{LoCoMo} and \textsc{LongMemEval}-S, budgeted completion alone raises annotated-evidence reach over one-shot ranking, and the full two-stage framework further improves evidence coverage and judged answer quality over same-store baselines. Ablations attribute most of the gain to broader evidence access rather than adaptive state control, and human checks show that gold-free sufficiency estimates remain too unreliable to serve as stopping criteria.
More broadly, these results motivate a different objective for long-term memory retrieval: not to rank individually relevant memories, but to construct a set whose members collectively satisfy the information requirements of the question. From this view, retrieval involves both \emph{composition}, assembling complementary support within reach, and \emph{completion}, continuing beyond the initial retrieval boundary when necessary. This makes set-level sufficiency a natural bridge between memory access and downstream reasoning: retrieval should deliver an answer-supporting evidentiary record, not simply a list of similar items. As agent histories become longer and required evidence is increasingly distributed across sessions, we expect this distinction between relevance and sufficiency to become central to how long-term memory systems are designed and evaluated.

\clearpage
\subsection*{AI use statement}
Generative AI systems were used in two roles.  As part of the reported method and
evaluation, language models perform requirement decomposition, candidate--requirement
coverage judgments, answer synthesis, answer judging, and blinded LLM-Suff@Set labeling
and adjudication; the corresponding roles, prompts, and evaluation rules are
documented in the paper and appendix.  Cursor and OpenAI Codex also assisted with
experimental planning, code generation and debugging, result organization,
literature discovery, and manuscript drafting and language editing.  The authors
inspected AI-produced code and outputs, reran the reported computations, verified
citations and numerical claims against stored experiment artifacts, and made all
final methodological and editorial decisions.  Public benchmark conversations,
questions, and reference answers were not generated or altered by these tools.  The
authors take responsibility for the complete contents of the submission.

\bibliography{references}
\bibliographystyle{iclr2027_conference}

\newpage

\appendix
\tableofcontents
\clearpage

\section{Evidence-law mapping}
\label{app:pardo}

Pardo asks what makes evidence sufficient, and answers by separating two judgments
that legal fact-finding already treats as distinct~\citep{pardo2023evidence}.
\emph{Relevance} is a property of an item: an exhibit is relevant when it has a
bearing on a fact of consequence.
\emph{Sufficiency} is a property of a record: the body of evidence is sufficient
when a reasonable finder of fact could make the finding from that record as
assembled.
The finder does not score items independently and then threshold the sum.
The question is whether the assembled exhibits, taken together, could support one
explanation of the disputed fact over its competitors.
Pardo therefore emphasizes \emph{explanatory facts}: relations between exhibits and
the elements of a finding, rather than a per-item probability that an exhibit is
``about'' the claim.

Memory QA uses the same two objects.
A memory $m\in\mathcal{M}$ is an exhibit, and $R(q,m)$ denotes its item-wise
relevance to question $q$: does this memory have any tendency to matter for $q$?
The retrieved set $E$ is the evidentiary record.
The ideal property $S(q,E)$ in \eqref{eq:sufficiency} adapts Pardo's
reasonable-finder test to conversational memory; the blinded LLM label
$\widehat S_{\mathrm{LLM}}(q,E)$ in \eqref{eq:llm_suffset} is its
reference-conditioned operational measurement.
The shutter-release case in \figref{fig:overview} is exactly this split: the
February 5 order is a relevant exhibit, and the photography memories around it are
also relevant, but the record cannot support the finding ``five days'' until the
February 10 arrival enters $E$.

The method in \Secref{sec:method} implements the two operations that this split
requires.
Information requirements $\mathcal{R}_q=\{r_j\}$ are the explanatory elements of
the finding: order date and arrival date in the running case; a fact and a bridge
in multi-hop questions; the items of a list.
The requirement-conditioned candidate assessment in Stage~I concerns relations
between an exhibit and a needed element, rather than a single query--memory score.
Stage~I composition is the assembly of a record whose exhibits jointly cover those
elements.
Stage~II completion is what a finder does when the current record is insufficient:
further exhibits are acquired, from the same store, rather than re-ranking exhibits
already known to be relevant.
The exclusion $K_t\subseteq\mathcal{M}\setminus E_t$ in \eqref{eq:update} is the
legal analogue of not re-admitting the same exhibit.
Gold Hit $G$ and Turn Hit ask whether an annotated answer-bearing exhibit has
entered the record; they correspond to coverage of at least one marked element,
which is why \Secref{sec:suff_eval} treats them as proxies associated with $S$ rather
than as the sufficiency test itself.

This mapping is conceptual, not a claim that agent memory should mimic courtroom
procedure.
It specifies the object we optimize: a record that could support a reasonable
finding, constructed by composing complementary exhibits and completing the record
when it cannot.

\section{Case trace for \figref{fig:overview}}
\label{app:case}

\Figref{fig:overview} uses question \texttt{b3c15d39} from
\textsc{LongMemEval}-S.
The answer requires two user turns: the remote shutter release was ordered on
February 5 and arrived on February 10.
Both lexical top-$k$ and \fca{} reach the order turn but miss the arrival turn,
leaving evidence recall at $50\%$ and producing an incorrect answer.
\mvpolicy{} reaches both turns and answers ``5 days'' under the same extractive
answerer and judge.

\begin{table}[h]
\centering
\caption{\textbf{Trace behind the relevance--sufficiency case.}
Case evidence recall counts the two required turns; the first two sets contain
relevant photography memories but miss one required date.}
\label{tab:case_trace}
\begin{tabular}{lcccc}
\toprule
Access & Order date & Arrival date & Gold recall & Judge \\
\midrule
Lexical top-$k$ & \checkmark & --- & 50\% & Wrong \\
\fca{} (Stage I) & \checkmark & --- & 50\% & Wrong \\
\mvpolicy{} & \checkmark & \checkmark & 100\% & Correct \\
\bottomrule
\end{tabular}
\end{table}

\section{Structured memory boundary}
\label{app:graph}

This appendix reports the cross-system comparison that was removed from the main
\textsc{LoCoMo} table.
The slice is the conv-30 overlap ($n{=}81$) under a unified extractive answerer and
judge.
BFR-Text and MRAgent-Flat still read flat turns; CoM uses an organized candidate
pool; Full-MRAgent uses a Cue--Tag--Content graph.
Open Domain is absent from this slice.
The comparison is a boundary, not a fair same-store baseline: the store, derived
facts, and tools move together.

\begin{table*}[t]
\centering
\setlength{\tabcolsep}{3.2pt}
\caption{\textbf{Cross-system boundary on the \textsc{LoCoMo} conv-30 overlap}
($n{=}81$), unified extractive answerer and judge; $\dagger$ marks rows re-scored
under this protocol.
Bold / underline: best / second-best.
Open Domain is absent from the slice.}
\label{tab:locomo_cross}
\begin{tabular}{l cc cc cc cc}
\toprule
 & \multicolumn{2}{c}{Multi-Hop} & \multicolumn{2}{c}{Temporal} & \multicolumn{2}{c}{Single-Hop} & \multicolumn{2}{c}{Overall} \\
\cmidrule(lr){2-3}\cmidrule(lr){4-5}\cmidrule(lr){6-7}\cmidrule(lr){8-9}
Method & F1 & J & F1 & J & F1 & J & F1 & J \\
\midrule
  BFR-Text$^\dagger$ & \textbf{41.4} & 54.5 & \textbf{100.0} & \textbf{100.0} & \underline{75.5} & 77.3 & \textbf{78.7} & 81.5 \\
  CoM$^\dagger$~\citep{xu2026com} & \underline{41.2} & \underline{63.6} & \underline{96.2} & \underline{96.2} & 71.8 & \underline{84.1} & 75.5 & \underline{85.2} \\
  MRAgent-Flat$^\dagger$~\citep{ji2026mragent} & 12.9 & 45.5 & 92.3 & 92.3 & 62.3 & 68.2 & 65.2 & 72.8 \\
  Full-MRAgent$^\dagger$~\citep{ji2026mragent} & 26.5 & \textbf{90.9} & 92.3 & \underline{96.2} & \textbf{78.8} & \textbf{95.5} & \underline{76.1} & \textbf{95.1} \\
\bottomrule
\end{tabular}
\end{table*}

\Tabref{tab:graph_confound} reports the same slice by store type.
Full-MRAgent reaches $88.9\%$ Gold Hit and $95.1\%$ judged accuracy,
while MRAgent-Flat reaches $67.9\%$ / $72.8\%$ and \textpolicy{} reaches
$77.8\%$ / $81.5\%$.
CoM reaches $80.2\%$ / $85.2\%$ under the same unified answerer.

\begin{table}[t]
\centering
\caption{\textbf{Flat acquisition vs.\ structured memory} on the $81$-question
overlap slice under a shared extractive answerer and judge.
Full MRAgent and CoM change the store and are reported as a boundary, not as
fair same-substrate baselines.}
\label{tab:graph_confound}
\begin{tabular}{lccc}
\toprule
Configuration & Store & Gold Hit (\%) & Judge Acc (\%) \\
\midrule
MRAgent-Flat & flat turns & 67.9 & 72.8 \\
\textpolicy{} & flat turns & 77.8 & 81.5 \\
\midrule
CoM & organized pool & 80.2 & 85.2 \\
Full MRAgent & Cue--Tag--Content graph & \textbf{88.9} & \textbf{95.1} \\
\bottomrule
\end{tabular}
\end{table}

Paired against \textpolicy{}, Full-MRAgent reaches gold on $12$ questions that
\textpolicy{} misses; the reverse occurs on $3$ questions.
Comparing Full-MRAgent with MRAgent-Flat, the full pipeline is judged correct alone
on $20$ questions, while the flat variant is correct alone on $2$.
Mean evidence-set sizes are $2.4$ (Full), $14.2$ (Flat), and $34.0$
(\textpolicy{}), so the graph pipeline reaches gold more often while retaining far
less text.
This pattern is consistent with associative precision rather than search volume
alone, but the store, tools, and derived facts still move together.
A causal graph-edge claim would require toggling the graph under identical tools.

\section{Algorithmic specification}
\label{app:algorithms}

\Algref{alg:fca_ms} specifies how Stage~I constructs the initial evidence set, and
\algref{alg:bfr_completion} gives the Stage~II completion loop shared by the fixed
access schedules.

\begin{algorithm}[H]
\caption{Stage I: Formal Concept Analysis for Memory Selection (FCA-MS)}
\label{alg:fca_ms}
\begin{algorithmic}[1]
\Require Question $q$, agent-memory notes $\widetilde{\mathcal M}$, source store
$\mathcal M$, usual retrieval cutoff $k$, expanded-pool size $N>k$,
prefilter cap $P$, grounding limit $b$, fallback size $f$
\Ensure Source-grounded initial evidence set $E_0$
\State $\mathcal C_q \gets
\operatorname{TopN}(\operatorname{Retrieve}_{\mathrm{agent}}(q,\widetilde{\mathcal M}),N)$
\Comment{Search beyond the usual top-$k$}
\State $\mathcal R_q \gets \operatorname{Decompose}(q)$
\Comment{Identify facts, bridges, list items, and temporal constraints}
\State $\mathcal P_q\gets\operatorname{FCAPrefilter}(\mathcal R_q,\mathcal C_q,P)$
\Comment{Reduce the expanded pool to at most $P$ notes}
\State $(\widehat S,\Gamma)\gets\operatorname{LLMMinSubset}(q,\mathcal P_q)$
\Comment{Request selected IDs and requirement coverage}
\State $S\gets\operatorname{Sanitize}(\widehat S,\Gamma,\mathcal P_q)$
\Comment{Keep valid IDs once; use coverage order if needed}
\If{$S=\varnothing$}
    \State $S\gets\operatorname{First}_f(\mathcal P_q)$
    \Comment{Fall back to the prefilter head}
\EndIf
\State $E_0\gets\operatorname{PositiveOverlapTop}_b(S,\mathcal M)$
\Comment{Score turns; keep positive scores in stable order}
\State \Return $E_0$
\Comment{Pass the composed evidence set to Stage II}
\end{algorithmic}
\end{algorithm}

\begin{algorithm}[H]
\caption{Stage II: Budgeted Flat Reconstruction (BFR) Completion}
\label{alg:bfr_completion}
\begin{algorithmic}[1]
\Require Question $q$, initial evidence $E_0$, flat store $\mathcal M$, schedule
$\pi\in\{\text{BFR-Text},\text{BFR-MV}\}$, budget
$B=(B_{\mathrm{step}},B_{\mathrm{call}},B_{\mathrm{new}})$, keep limit $k$
\Ensure Completed evidence set $E_T$
\State $E\gets E_0$; $t\gets0$; $b_{\mathrm{call}}\gets0$;
$b_{\mathrm{new}}\gets0$
\Comment{Initialize the cumulative evidence state and counters}
\While{$t<B_{\mathrm{step}}$ and $b_{\mathrm{call}}<B_{\mathrm{call}}$
and $b_{\mathrm{new}}<B_{\mathrm{new}}$}
    \If{$\pi=\text{BFR-Text}$}
        \State $A_t\gets\{(\mathrm{text},q)\}$
        \Comment{Advance beyond the current text-ranking cutoff}
    \Else
        \State $A_t\gets\{(\mathrm{text},q),
        (\mathrm{entity},\operatorname{FirstPerson}(q)),
        (\mathrm{session},\operatorname{Latest}(E))\}$
        \Comment{Diversify access through text, entity, and session cues}
    \EndIf
    \State $C_t\gets\varnothing$
    \Comment{Hold the current evidence fixed throughout this round}
    \ForAll{$(v,c)\in A_t$ in schedule order}
        \If{$c=\varnothing$ or $b_{\mathrm{call}}=B_{\mathrm{call}}$}
            \State \textbf{continue}
            \Comment{Skip unavailable cues or exhausted call budgets}
        \EndIf
        \State $C_t\gets C_t\cup R_v(c,\mathcal M\setminus E)$
        \State $b_{\mathrm{call}}\gets b_{\mathrm{call}}+1$
        \Comment{Collect candidates relative to the same round-start evidence}
    \EndFor
    \State $C_t\gets\operatorname{Unique}(C_t)$
    \State $K_t\gets\kappa_\pi(C_t,q,E,k)$
    \Comment{Apply the retention rule once to the merged candidates}
    \State $K_t\gets\operatorname{ClipToBudget}
        (K_t,B_{\mathrm{new}}-b_{\mathrm{new}})$
    \State $E\gets E\cup K_t$;
        $b_{\mathrm{new}}\gets b_{\mathrm{new}}+|K_t|$
    \Comment{Update evidence after the round's candidate union and retention}
    \State $t\gets t+1$
    \Comment{Proceed until an observable budget is exhausted}
\EndWhile
\State \Return $E_T\gets E$
\Comment{Supply the completed set to the fixed answerer}
\end{algorithmic}
\end{algorithm}

\section{Implementation details}
\label{app:impl}

\paragraph{Same-store baselines.}
Within each benchmark, the flat conversational turn pool, extractive answerer,
and Mem0-style judge are fixed across arms.
Mem0 uses BM25-lite over the turn pool; A-Mem uses dense top-$k$ over the same
pool without note evolution.
CoM$^*$ uses MiniLM candidates with anchor, chain, and blocking.
MRAgent-Flat$^*$ restricts its tools to flat turns, omitting the
Cue--Tag--Content graph and derived facts.
The starred methods are controlled same-store adaptations, not the official
full systems.

\paragraph{\fca{} initialization.}
In the \textsc{LoCoMo} pilot, dense A-Mem note retrieval yields $N=50$
candidates.  \texttt{fca\_prefilter\_candidates} uses the decomposed requirements
to retain at most $P=20$ notes.  The \texttt{setr\_pool} v1 prompt asks an LLM
for the smallest subset that jointly answers the question, with no fixed
note-count cap.  Invalid or duplicate IDs are removed; an empty selection
uses requirement-coverage output and then the prefilter head as fallback.
The resulting sets contain $1$--$5$ notes (mean $1.86$).  Only these notes
enter \texttt{map\_amem\_ids\_to\_turns(..., top\_n=10)}, which ranks source
turns by positive token overlap with the selected note content and context.
All $200$ pilot questions yielded $10$ mapped turns; these turn IDs form $E_0$
for Stage~II.  A deployment with direct source pointers can replace this
approximate alignment with an exact grounding map $\phi$.

\paragraph{Budgeted completion.}
\textpolicy{} is recorded as \texttt{IterRAG-Text} in experiment logs.
It resubmits the original question to the text index, excludes current evidence ids,
and retains the first eight unseen results.
\mvpolicy{} is recorded as \texttt{Q-Fixed}; it runs the fixed text$\to$entity$\to$session
schedule in \Secref{sec:bfr_policies}.
The default $B_{\mathrm{mid}}$ point allows $3$ rounds, $6$ retrieval calls, and $30$
newly retained records.  This repeated, exclusion-aware schedule generalizes
one-pass acquisition by continuing to collect unseen support until the observable
budget is exhausted.

\section{Data usage and sealing}
\label{app:sealing}

All $10$ LoCoMo conversations were visible during development, so every LoCoMo number
measures stability across conversations rather than generalization to unseen data.
LongMemEval-S was not read, scored, or inspected until the LoCoMo protocol, arm
definitions, budget grid, and metrics were frozen.
After the initial BFR evaluation on \textsc{LongMemEval}-S,
CoM$^*$ and MRAgent-Flat$^*$ were evaluated under the same
flat-store protocol and included in the main comparison.

\section{Dataset provenance and derived evaluation sets}
\label{app:data_provenance}

\paragraph{Public source benchmarks.}
\textsc{LoCoMo} was introduced by \citet{maharana2024evaluating} as a benchmark for
very long-term conversational memory.
Its released QA corpus contains ten long, multi-session conversations produced by
the authors' persona- and event-conditioned conversation-generation framework.
The release pairs questions and reference answers with category labels and, when
available, the dialogue-turn identifiers that support the answer.
We use the released conversations and QA annotations; we do not generate new
dialogue content or alter the reference answers.

\textsc{LongMemEval} was introduced by \citet{DBLP:conf/iclr/WuWYZCY25} using an
attribute-controlled pipeline that places evidence sessions inside coherent,
timestamped conversation histories.
We use the official cleaned \textsc{LongMemEval}-S file, which contains $500$
questions covering single-session user and assistant facts, preferences,
multi-session reasoning, temporal reasoning, and knowledge updates.
The complete $500$-question split is evaluated.
Turn Hit and turn recall use the official answer-bearing-turn fields;
$479$ questions contain at least one such turn.
The main Turn Hit results use all $500$ questions and count the $21$ questions
without an annotated answer turn as misses.
The matched-C3 analysis in \appref{app:mv-text} instead reports Turn Hit
over the $479$ questions with nonempty answer-turn annotations.

\paragraph{Derived LoCoMo evaluation sets.}
The $200$-question pilot is a fixed, stratified subset drawn from the official ten
conversations and shared by every same-store arm.
The pilot was sampled with seed $20260908$ and contains
33 Multi-Hop, 50 Temporal, 8 Open Domain, and 109 Single-Hop
questions, using the released manifest labels without reclassification.
It is a development subset for controlled method, ablation, annotation, and budget
comparisons, not an official held-out split.
The $1{,}536$-question gold subset contains all questions in our LoCoMo release with
usable supporting dialogue-turn identifiers; it is used only for development
robustness across conversations.
The conv-30 slice contains the $81$ question ids with aligned outputs for the
cross-system comparison in \appref{app:graph}.
It is an overlap set rather than an independently sampled benchmark, and Open Domain
is absent from this slice.

\paragraph{Evidence-state and budget artifacts.}
For method-aligned sufficiency analysis, seven arms (Mem0, A-Mem, CoM$^*$,
MRAgent-Flat$^*$, FCA-MS, BFR-Text C3, and BFR-MV C6) are evaluated on each
of the fixed $200$ questions
by the same blinded LLM-Suff@Set protocol in \appref{app:annotation}.
These $1{,}400$ method--question evaluations concern retrieval outputs rather than
additional conversations.  The budget grid applies
the fixed BFR access policies and call caps to the same pilot; identical evidence
states are deduplicated by question and evidence hash.
These artifacts evaluate retrieval behavior over the public benchmark; they are not
presented as new datasets.

\begin{table}[H]
\centering
\setlength{\tabcolsep}{3.5pt}
\caption{\textbf{Provenance of the public benchmarks and derived evaluation sets.}
Only LoCoMo and LongMemEval-S are independently released benchmarks; all other rows
are fixed subsets or evidence states constructed from them.}
\label{tab:data_provenance}
\begin{tabular}{>{\raggedright\arraybackslash}p{0.20\linewidth}
                >{\raggedright\arraybackslash}p{0.13\linewidth}
                >{\raggedright\arraybackslash}p{0.13\linewidth}
                >{\raggedright\arraybackslash}p{0.43\linewidth}}
\toprule
Evaluation set & Parent & Size & Construction and role \\
\midrule
LoCoMo release & Public benchmark & 10 conversations & Official multi-session conversations, QA pairs, categories, and supporting turn ids. \\
LoCoMo pilot & LoCoMo & 200 questions & Fixed stratified development subset shared by all same-store arms. \\
LoCoMo gold subset & LoCoMo & 1,536 questions & All questions with usable gold turn ids; used for development robustness. \\
conv-30 overlap & LoCoMo & 81 questions & Qid intersection for cross-system scoring; not a held-out split. \\
LLM-Suff@Set evaluations & LoCoMo pilot & 1,400 arm--question pairs & $200$ questions paired with Mem0, A-Mem, CoM$^*$, MRAgent-Flat$^*$, FCA-MS, BFR-Text C3, and BFR-MV C6 under blinded method identities. \\
Budget grid & LoCoMo pilot & Fixed policy--budget runs & BFR-Text and BFR-MV call budgets evaluated before evidence-hash deduplication. \\
LongMemEval-S & Public benchmark & 500 questions & Official cleaned small split, evaluated in full after protocol freezing. \\
\bottomrule
\end{tabular}
\end{table}

\section{Same-store \textsc{LoCoMo} controls}
\label{app:locomo_controls}

\begin{table}[H]
\centering
\caption{\textbf{Retrieval cost of the adapted same-store controls} on the fixed
$200$-question pilot.  Calls and evidence-set size are realized means.}
\label{tab:flat_control_cost}
\begin{tabular}{lccc}
\toprule
Method & Gold Hit & Calls & Evidence-set size \\
\midrule
CoM$^*$ & 49.0 & 1.00 & 10.0 \\
MRAgent-Flat$^*$ & 62.5 & 2.54 & 11.8 \\
\bottomrule
\end{tabular}
\end{table}

\begin{table}[H]
\centering
\setlength{\tabcolsep}{4pt}
\caption{\textbf{Paired LoCoMo comparisons against the adapted controls.}
$\Delta$ is the first method minus the second.  Judge uses two-sided exact McNemar;
Stem F1 uses a 10,000-sample question-level paired bootstrap (seed $20260908$).}
\label{tab:locomo_pairwise}
\begin{tabular}{lrrrr}
\toprule
Comparison & $\Delta J$ & $p$ & $\Delta$ F1 & 95\% CI \\
\midrule
BFR-Text $-$ CoM$^*$ & $+16.0$ & $1.4\times10^{-5}$ & $+17.8$ & $[11.3,24.2]$ \\
BFR-MV $-$ CoM$^*$ & $+16.5$ & $8.7\times10^{-6}$ & $+18.2$ & $[11.8,24.8]$ \\
BFR-Text $-$ MRAgent-Flat$^*$ & $+5.5$ & $0.071$ & $+9.8$ & $[5.0,14.8]$ \\
BFR-MV $-$ MRAgent-Flat$^*$ & $+6.0$ & $0.043$ & $+10.2$ & $[5.5,15.2]$ \\
\bottomrule
\end{tabular}
\end{table}

CoM$^*$ completes all $200$ questions.  MRAgent-Flat$^*$ completes $199$;
\texttt{q\_00208} has empty evidence and remains in the denominator as an empty
prediction.  These rows are distinct from the $81$-question full-system comparison
in \appref{app:graph}.

\section{Budget sweep}
\label{app:budget}

\begin{table}[H]
\centering
\setlength{\tabcolsep}{3.2pt}
\caption{\textbf{BFR access schedules by question type on \textsc{LoCoMo}}
($n{=}200$).  F1 and J are percentages; Text and MV use the C3 and C6
operating points, respectively.}
\label{tab:locomo_schedule_types}
\begin{tabular}{l cc cc cc cc cc}
\toprule
 & \multicolumn{2}{c}{Multi-Hop} & \multicolumn{2}{c}{Temporal} &
 \multicolumn{2}{c}{Open Domain} & \multicolumn{2}{c}{Single-Hop} &
 \multicolumn{2}{c}{Overall} \\
\cmidrule(lr){2-3}\cmidrule(lr){4-5}\cmidrule(lr){6-7}
\cmidrule(lr){8-9}\cmidrule(lr){10-11}
Schedule & F1 & J & F1 & J & F1 & J & F1 & J & F1 & J \\
\midrule
BFR-Text & 44.7 & 54.5 & 78.1 & 80.0 & 1.9 & 0.0 &
74.9 & 75.2 & 67.8 & 70.0 \\
BFR-MV & 44.6 & 48.5 & 78.1 & 80.0 & 2.7 & 12.5 &
75.7 & 77.1 & 68.3 & 70.5 \\
\bottomrule
\end{tabular}
\end{table}

\begin{table}[H]
\centering
\setlength{\tabcolsep}{4pt}
\caption{\textbf{Evidence Recall and Complete Evidence Recall by question type on
the \textsc{LoCoMo} pilot.}
Text and MV use the C3 and C6 operating points, respectively.}
\label{tab:budget_type_recall}
\begin{tabular}{l r cc cc cc}
\toprule
& & \multicolumn{2}{c}{FCA-MS} & \multicolumn{2}{c}{BFR-Text} &
\multicolumn{2}{c}{BFR-MV} \\
\cmidrule(lr){3-4}\cmidrule(lr){5-6}\cmidrule(lr){7-8}
Question type & $n$ & ER & CER & ER & CER & ER & CER \\
\midrule
Multi-Hop   & 33  & 24.8 & 6.1 & \textbf{42.6} & \textbf{18.2} & 41.8 & 12.1 \\
Temporal    & 50  & 68.0 & 66.0 & \textbf{82.0} & \textbf{82.0} & \textbf{82.0} & \textbf{82.0} \\
Open Domain & 8   & 18.8 & 12.5 & 27.1 & 12.5 & \textbf{31.2} & \textbf{25.0} \\
Single-Hop  & 109 & 42.7 & 41.3 & \textbf{72.0} & \textbf{71.6} & 70.6 & 70.6 \\
\midrule
Overall     & 200 & 45.1 & 40.5 & \textbf{67.9} & \textbf{63.0} & 67.2 & 62.0 \\
\bottomrule
\end{tabular}
\end{table}

\begin{table}[H]
\centering
\setlength{\tabcolsep}{4pt}
\caption{\textbf{Blinded LLM-Suff@Set by question type on the \textsc{LoCoMo}
pilot.}  Differences use BFR minus FCA-MS.  Open Domain contains only eight
questions.}
\label{tab:budget_type_suff}
\begin{tabular}{l r ccc cc}
\toprule
Type & $n$ & FCA & Text & MV & $\Delta$Text & $\Delta$MV \\
\midrule
Multi-Hop   & 33  & 15.2 & 27.3 & \textbf{30.3} & $+12.1$ & $+15.2$ \\
Temporal    & 50  & 60.0 & \textbf{74.0} & \textbf{74.0} & $+14.0$ & $+14.0$ \\
Open Domain & 8   & 0.0 & 0.0 & 0.0 & $0.0$ & $0.0$ \\
Single-Hop  & 109 & 45.9 & 71.6 & \textbf{73.4} & $+25.7$ & $+27.5$ \\
\midrule
Overall     & 200 & 42.5 & 62.0 & \textbf{63.5} & $+19.5$ & $+21.0$ \\
\bottomrule
\end{tabular}
\end{table}

Overall, BFR-Text $-$ FCA-MS is $+4.0$ points ($p=0.332$), while
BFR-MV $-$ FCA-MS is $0.0$ points ($p=1.000$).  No type-specific LLM-Suff@Set
difference is significant after Holm correction.

\begin{table}[H]
\centering
\setlength{\tabcolsep}{3.2pt}
\caption{\textbf{Matched-budget quality on the \textsc{LoCoMo} pilot}
($n{=}200$). Calls and set size are realized means; size counts grounded source
turns. FCA-MS selects $1$--$5$ notes (mean $1.86$) before mapping to $10$
turns. $J$: judged accuracy; CER: Complete Evidence Recall.}
\label{tab:budget}
\begin{tabular}{llrrrrrr}
\toprule
Method & Budget & Calls & Size & F1 & $J$ & \shortstack{LLM-\\Suff@Set} & CER \\
\midrule
FCA-MS & C0 & 0.00 & 10.00 & 48.6 & 53.5 & 42.5 & 40.5 \\
\midrule
\multirow{6}{*}{BFR-MV}
 & C1 & 1.00 & 21.91 & 65.4 & 67.5 & 57.5 & 59.0 \\
 & C2 & 2.00 & 29.05 & 66.8 & 68.5 & 58.0 & 59.5 \\
 & C3 & 3.00 & 29.84 & 66.8 & 69.5 & 58.0 & 59.5 \\
 & C4 & 4.00 & 39.72 & 68.3 & 71.5 & 61.0 & 62.5 \\
 & C6 & 5.99 & 39.88 & 68.3 & 70.5 & 63.5 & 62.0 \\
 & C9 & 8.98 & 59.23 & 69.9 & 72.0 & 62.5 & 64.0 \\
\midrule
\multirow{4}{*}{BFR-Text}
 & C1 & 1.00 & 18.00 & 64.4 & 67.5 & 56.0 & 57.0 \\
 & C2 & 2.00 & 26.00 & 66.3 & 69.0 & 59.5 & 60.5 \\
 & C3 & 3.00 & 34.00 & 67.8 & 70.0 & 62.0 & 63.0 \\
 & C4/C6 & 4.00 & 40.00 & 69.3 & 70.0 & 62.5 & 63.5 \\
\bottomrule
\end{tabular}
\end{table}

\tabref{tab:budget} gives the full operating-point grid discussed in \Secref{sec:budget_question}.
Calls are realized means rather than caps.
Across the full grid, BFR-Text occupies the three-call frontier for Judge, F1,
LLM-Suff@Set, and Complete Evidence Recall, while BFR-MV reaches the highest Judge score
at C9 and contributes non-dominated Judge points at C4 and C9.  These operating
points make the acquisition-cost frontier explicit.  The source-level Pareto
classification uses realized calls and evidence-set size and treats higher quality
and lower cost as preferred.
\tabref{tab:motivation_decomp} contrasts the Stage-I evidence set with the
completed text and multi-view schedules on both benchmarks.
\tabref{tab:ablation} reports the state-conditioned controllers from
\Secref{sec:bfr_policies}; their results motivate the simpler fixed completion
schedules used in the main framework.

\begin{table}[H]
\centering
\caption{\textbf{Annotated-evidence reach before and after completion.}
LoCoMo reports Gold Hit and LongMemEval-S reports Turn Hit.  The one-shot
LongMemEval-S cell lists FCA-MS / Mem0; parenthesized gains use FCA-MS.}
\label{tab:motivation_decomp}
\begin{tabular}{lcc}
\toprule
Step & \textsc{LoCoMo} Gold Hit & \textsc{LME}-S Turn Hit \\
\midrule
One-shot (\fca{} / Mem0) & 51.0 & 85.0 / 64.8 \\
BFR-MV & 73.5 (+22.5) & 91.4 (+6.4) \\
\midrule
BFR-Text (vs.\ one-shot) & 72.5 (+21.5) & 88.6 (+3.6) \\
\bottomrule
\end{tabular}
\end{table}

\begin{table}[H]
\centering
\setlength{\tabcolsep}{3pt}
\caption{\textbf{Evidence-state ablations} on the \textsc{LoCoMo} pilot at $B_{\mathrm{mid}}$.
Brackets: clustered $95\%$ CI. Budgeted access drives the gain; state control does not.
CI for Text vs.\ Initial $[19.1,26.7]$; State vs.\ MV $[-0.8,4.5]$.}
\label{tab:ablation}
\begin{tabular}{llcc}
\toprule
Factor & Variant & $G$ (\%) & $\Delta$ \\
\midrule
\multirow{2}{*}{Access}
 & Initial & 51.0 & ref. \\
 & \textpolicy{} & \textbf{72.5} & $+21.5$ \\
\midrule
\multirow{3}{*}{State}
 & \mvpolicy{} & 73.5 & ref. \\
 & \statepolicy{} & 73.0 & $-0.5$ \\
 & \adaptivepolicy{} & 73.0 & $-0.5$ \\
\bottomrule
\end{tabular}
\end{table}

\statepolicy{} adds a missing-requirement cue and coverage/bridge retention;
\adaptivepolicy{} also selects views and stops early from the evidence state.
At $B_{\mathrm{mid}}$ on \textsc{LoCoMo}, both reach $73.0\%$ Gold Hit versus
$73.5\%$ for fixed \mvpolicy{}.  The clustered $95\%$ CI for
\statepolicy{} minus \mvpolicy{} is $[-0.8,4.5]$ points, so this ablation does
not establish a gain over fixed access.

\section{Development robustness}
\label{app:5fold}

On the broader development-contact \textsc{LoCoMo} set ($n{=}1536$),
conversation-level five-fold analysis raises Gold Hit from $54.8\%$ for initial
retrieval to $78.0\%$ and $78.1\%$ for text and multi-view completion.

\begin{table}[H]
\centering
\caption{\textbf{Development robustness on all \textsc{LoCoMo} questions with gold ids}
($n{=}1536$). Conversation-level $5$-fold; all conversations had development contact.}
\label{tab:5fold_app}
\begin{tabular}{lc}
\toprule
Method & Gold Hit (\%) $\uparrow$ \\
\midrule
Initial retrieval & 54.8 \\
\textpolicy{} & 78.0 \\
\mvpolicy{} & 78.1 \\
\bottomrule
\end{tabular}
\end{table}

\section{Full LongMemEval-S type breakdown}
\label{app:lme_types}

The six released types show the multi-session and temporal retrieval-coverage
pattern behind the answer-quality results in \tabref{tab:main_lme}.

\begin{table}[H]
\centering
\caption{\textbf{Full type breakdown under the unified flat-store protocol on
\textsc{LongMemEval}-S} (all $500$ questions; unannotated answer-turn items count as misses).
Upd.: knowledge-update; Mul.: multi-session; Tmp.: temporal; Pref.: preference;
Asst./User: single-session. The six type columns and $G$ report Turn Hit;
$J$ is judged accuracy over all questions.}
\label{tab:lme_types_full}
\begin{tabular}{lcccccccc}
\toprule
Method & Upd. & Mul. & Tmp. & Pref. & Asst. & User & $G$ & $J$ \\
\midrule
Mem0 & 78.2 & 53.4 & 61.7 & 16.7 & 87.5 & 80.0 & 64.8 & 72.4 \\
A-Mem & 88.5 & 82.0 & 75.2 & 70.0 & 98.2 & 81.4 & 82.2 & 68.6 \\
CoM & 88.5 & 81.2 & 72.9 & 63.3 & 98.2 & 78.6 & 80.6 & 68.6 \\
MRAgent-Flat & 79.5 & 75.9 & 78.9 & 60.0 & 91.1 & 81.4 & 78.8 & 71.4 \\
\fca{} & 89.7 & 84.2 & 80.5 & 70.0 & 100.0 & 84.3 & 85.0 & 70.4 \\
\textpolicy{} & 91.0 & 85.7 & 88.7 & 70.0 & 100.0 & 90.0 & 88.6 & 81.6 \\
\mvpolicy{} & 92.3 & 91.0 & 91.0 & 76.7 & 100.0 & 91.4 & 91.4 & 82.2 \\
\bottomrule
\end{tabular}
\end{table}

\section{Comparison with learned residual set completion}
\label{app:rsc}

\begin{table}[H]
\centering
\caption{\textbf{Learned residual set completion (RSC) comparison} on the
\textsc{LoCoMo} pilot ($n{=}200$).  The three completion policies use a matched
budget of three retrieval calls; initial retrieval is shown as a zero-call reference.}
\label{tab:rsc_neg}
\begin{tabular}{lcc}
\toprule
Method & Gold Hit (\%) $\uparrow$ & Calls \\
\midrule
Initial retrieval & 51.0 & 0.00 \\
\textpolicy{} C3 & \textbf{72.5} & 3.00 \\
\mvpolicy{} C3 & \underline{70.5} & 3.00 \\
RSC (dense residual) & 58.0 & 3.00 \\
\bottomrule
\end{tabular}
\end{table}

Under the same three-call budget, the fixed text and multi-view schedules recover
more annotated gold evidence than RSC: Gold Hit is $72.5\%$ for BFR-Text C3,
$70.5\%$ for BFR-MV C3, and $58.0\%$ for RSC.  Under the current training signal
and flat-store protocol, learning a residual query therefore does not improve over
the fixed completion schedules.

\section{Annotation protocol}
\label{app:annotation}

Suff@Set is labelled on a question--evidence snapshot rather than on the system
answer.  The blinded input contains the question, reference answer, and anonymized
evidence text; it withholds the method, budget, system answer, and gold evidence
identifiers.  Two independent judges (\texttt{gpt-5.5} and
\texttt{claude-sonnet-4-6}) assign \emph{sufficient}, \emph{insufficient}, or
\emph{uncertain}; \texttt{gpt-6-sol} adjudicates disagreements.  All calls use
temperature zero.  Because the protocol includes the reference answer and uses
model judges, Suff@Set is an LLM-only, reference-conditioned offline evaluation
measure rather than a human annotation or an inference-time stopping signal.
An uncertain or missing final label is excluded from the binary-rate denominator;
the originally reported FCA-MS, BFR-Text C3, and BFR-MV C6 pilot produced no
uncertain labels. No human Suff@Set labels are reported.

\section{Metric definitions}
\label{app:metrics}

The sufficiency-oriented measurements used in the main text are defined here
because they answer different questions about the retrieved set.

\paragraph{Set sufficiency $S$.}
$S(q,E)=1$ when a careful reader can determine $y^\star(q)$ from the question and
the retrieved evidence text alone (\eqref{eq:sufficiency}).
The blinded LLM protocol estimates this property through
$\widehat S_{\mathrm{LLM}}(q,E)$ (\eqref{eq:llm_suffset};
\appref{app:annotation}).
$S$ is a reference-conditioned record-level test; it is not computed from
similarity ranks or gold evidence identifiers.

\paragraph{Gold Hit $G$.}
Gold Hit is $1$ when the source identifiers in $E$ intersect the
annotated supporting-record identifiers (\eqref{eq:reach}).
$G$ answers a narrower question than $S$: did at least one marked support record
enter the context?
A hit can still leave a compositional question underspecified, so we do not treat
$G$ as sufficiency.

\paragraph{Evidence Recall and Complete Evidence Recall.}
For questions with a nonempty annotated gold set, Evidence Recall is the fraction
of gold identifiers in $\operatorname{src}(E)$; Complete Evidence Recall requires
all gold identifiers to be present (\eqref{eq:evidence_recall}).  The former measures
partial recovery; the latter requires all annotated support records.  Complete
Evidence Recall concerns completeness relative to benchmark annotations and is not
identical to semantic Suff@Set.

\paragraph{Turn Hit.}
Turn Hit is $1$ when the source identifiers in $E$ include any annotated
answer-bearing turn (\eqref{eq:turn_hit}).
Main \textsc{LongMemEval}-S results average over all $500$ questions,
counting unannotated items as misses; \appref{app:mv-text} separately uses
the $479$ questions with answer-turn annotations.
Like Gold Hit, it measures access to marked evidence rather than support of
the complete answer.

\paragraph{Answer scores.}
Stem F1 follows the LoCoMo / MRAgent normalization with Porter stemming.
Judge accuracy is the Mem0-style CORRECT/WRONG decision under
\texttt{gpt-5.5} at temperature $0$ (\appref{app:prompts}).
Native free-form scores from other systems are never mixed into same-store rankings
without the unified-answerer note.

\section{Prompts}
\label{app:prompts}

We list the prompts used in our evaluation pipeline.
Formatting follows the prompt-box style common in recent memory-agent papers.

\begin{promptbox}{LLM-Judge Prompt (Mem0-style)}
Your task is to label an answer to a question as 'CORRECT' or 'WRONG'. You will be given the following data:
\begin{itemize}[leftmargin=1.25em]
  \item a question (posed by one user to another user),
  \item a 'gold' (ground truth) answer,
  \item a generated answer
\end{itemize}
which you will score as CORRECT/WRONG.

The point of the question is to ask about something one user should know about the other user based on their prior conversations.
The gold answer will usually be a concise and short answer that includes the referenced topic.
The generated answer might be much longer, but you should be generous with your grading---as long as it touches on the same topic as the gold answer, it should be counted as CORRECT.

For time related questions, the gold answer will be a specific date, month, year, etc.
The generated answer might be much longer or use relative time references, but you should be generous with your grading---as long as it refers to the same date or time period as the gold answer, it should be counted as CORRECT.
Even if the format differs (e.g., ``May 7th'' vs ``7 May''), consider it CORRECT if it is the same date.

Now it's time for the real question:
\begin{verbatim}
Question: {question}
Gold answer: {gold_answer}
Generated answer: {generated_answer}
\end{verbatim}

First, provide a short (one sentence) explanation of your reasoning, then finish with CORRECT or WRONG.
Do NOT include both CORRECT and WRONG in your response.

Just return the label CORRECT or WRONG in a json format with the key as ``label'':
\begin{lstlisting}[style=json]
{"label": "CORRECT"}
\end{lstlisting}
\end{promptbox}

\begin{promptbox}{Suff@Set Judging Inputs and Decision Rule}
You will see a question, its reference answer, and an anonymized retained evidence
set from a conversational memory store.  The method, budget, system answer, and gold
memory identifiers are \textbf{not} shown.

\textbf{Task:} Decide whether the evidence text supports the complete reference answer.
\begin{itemize}[leftmargin=1.25em]
  \item Label \texttt{sufficient} if the complete answer is supported by the provided evidence.
  \item Label \texttt{insufficient} if a required fact, bridge, or temporal detail is missing.
  \item Label \texttt{uncertain} only when the evidence does not permit a reliable decision.
  \item Do not use outside knowledge beyond what is written in the evidence.
\end{itemize}
Two judges apply this rule independently; a third model adjudicates disagreements.
\end{promptbox}

\begin{promptbox}{Extractive Answerer Cue}
Given the question and the ordered evidence snippets, produce a short answer string that is supported by the evidence.
Prefer spans that appear in the evidence.
If multiple candidates exist, choose the most specific span that answers the question.
Do not invent entities that are absent from the evidence set.
\end{promptbox}

\begin{promptbox}{BFR Multi-View Schedule (deterministic)}
At each round $t$, issue up to three retrieval actions in fixed order:
\begin{enumerate}[leftmargin=1.25em]
  \item \textbf{text}: cue $=$ the original question $q$.
  \item \textbf{entity}: cue $=$ first person name extracted from $q$ (skip if none).
  \item \textbf{session}: cue $=$ most recently retained evidence record (skip if none).
\end{enumerate}
Exclude ids already in $E_t$. Retain policy-selected unseen candidates and union into $E_{t+1}$.
Stop when the call / round / retention budget is exhausted.
\end{promptbox}

\section{Matched three-call comparison of \mvpolicy{} and \textpolicy{}}
\label{app:mv-text}

Both policies start from the same \fca{} initialization and make 3.00 retrieval
calls. Differences are \mvpolicy{} minus \textpolicy{}.

\Tabref{tab:mv-locomo-gain} reports selected question-type results at
the matched three-call budget. \textpolicy{} has higher Temporal
LLM-Suff@Set; the Open Domain rows report the observed answer scores
on eight questions.

\begin{table}[H]
\centering
\setlength{\tabcolsep}{4pt}
\caption{\textbf{Selected question-type results at a matched three-call budget.}
\textsc{LoCoMo}, C3, $n{=}200$. Values are percentages; differences are
\mvpolicy{} minus \textpolicy{} in percentage points, computed before rounding.
Open Domain has eight questions.}
\label{tab:mv-locomo-gain}
\begin{tabular}{llrrrr}
\toprule
Question type & Metric & $n$ & BFR-Text & BFR-MV & $\Delta$ \\
\midrule
Open Domain & Judge & 8 & 0.0 & 12.5 & $+12.5$ \\
Temporal & LLM-Suff@Set & 50 & 74.0 & 70.0 & $-4.0$ \\
Open Domain & F1 & 8 & 1.9 & 3.4 & $+1.6$ \\
\bottomrule
\end{tabular}
\end{table}

\begin{figure}[H]
\begin{minipage}[t]{0.54\linewidth}
\vspace{0pt}
On \textsc{LongMemEval}-S the same budget favors \mvpolicy{} on Turn Hit
(\figref{fig:mv-lme-turn}). Overall is $95.2$ against $92.5$ ($+2.7$~pp) on
the $479$ items with an answer-turn label, whereas the main tables
average Turn Hit over all $500$ questions. Preference is $+6.7$ ($n{=}30$)
and multi-session is $+5.6$ ($n{=}125$); assistant questions are tied at
$100\%$. Paired outcomes are 13 to 0 ($p{=}2.4\times10^{-4}$).
Multi-session is 7 to 0 (raw $p{=}0.016$; Bonferroni over six types
$p{=}0.094$). Judge does not separate the two policies.
\end{minipage}\hfill
\begin{minipage}[t]{0.40\linewidth}
\vspace{0pt}
\includegraphics[width=\linewidth]{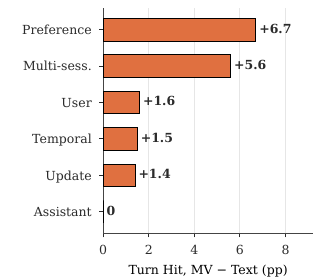}
\end{minipage}
\caption{\textbf{\textsc{LongMemEval}-S Turn Hit under matched C3 access.}
\mvpolicy{} minus \textpolicy{}, on $479$ items with an answer-turn label;
differences are percentage points.}
\label{fig:mv-lme-turn}
\end{figure}

\end{document}